\documentclass[conference]{IEEEtran}
\IEEEoverridecommandlockouts
\usepackage{cite}
\usepackage{graphicx}
\usepackage{picinpar}
\usepackage{stfloats}
\usepackage{url}
\usepackage{flushend}
\usepackage[latin1]{inputenc}
\usepackage{colortbl}
\usepackage{soul}
\usepackage{multirow}
\usepackage{pifont}
\usepackage{alltt}
\usepackage{enumerate}
\usepackage{siunitx}
\usepackage{pbox}
\usepackage{amsmath,amssymb,amsfonts}
\usepackage{amsthm}
\usepackage{algorithmic}
\usepackage{algorithm}
\usepackage{bm}
\usepackage{graphbox}
\usepackage{textcomp}
\usepackage{xcolor}
\usepackage[export]{adjustbox}
\usepackage{color}
\usepackage{tikz}
\usepackage{subcaption}
\usepackage{subfig}   

\def\BibTeX{{\rm B\kern-.05em{\sc i\kern-.025em b}\kern-.08em
    T\kern-.1667em\lower.7ex\hbox{E}\kern-.125emX}}

\newcommand\copyrighttext{%
  \footnotesize \textcopyright This paper has been accepted for presentation and publication at the Eight Iberian Robotics Conference, 2025. Please, when citing the paper, refer to the official manuscript.}
\newcommand\copyrightnotice{%
\begin{tikzpicture}[remember picture,overlay]
\node[anchor=south,yshift=10pt] at (current page.south) {\fbox{\parbox{\dimexpr\textwidth-\fboxsep-\fboxrule\relax}{\copyrighttext}}};
\end{tikzpicture}%
}

\begin{document}

\newcommand{\calA}{{\cal A}}
\newcommand{\calB}{{\cal B}}
\newcommand{\calC}{{\cal C}}
\newcommand{\calD}{{\cal D}}
\newcommand{\calE}{{\cal E}}
\newcommand{\calF}{{\cal F}}
\newcommand{\calG}{{\cal G}}
\newcommand{\calH}{{\cal H}}
\newcommand{\calI}{{\cal I}}
\newcommand{\calJ}{{\cal J}}
\newcommand{\calK}{{\cal K}}
\newcommand{\calL}{{\cal L}}
\newcommand{\calM}{{\cal M}}
\newcommand{\calN}{{\cal N}}
\newcommand{\calO}{{\cal O}}
\newcommand{\calP}{{\cal P}}
\newcommand{\calQ}{{\cal Q}}
\newcommand{\calR}{{\cal R}}
\newcommand{\calS}{{\cal S}}
\newcommand{\calT}{{\cal T}}
\newcommand{\calU}{{\cal U}}
\newcommand{\calV}{{\cal V}}
\newcommand{\calW}{{\cal W}}
\newcommand{\calX}{{\cal X}}
\newcommand{\calY}{{\cal Y}}
\newcommand{\calZ}{{\cal Z}}

\newcommand{\frakA}{{\mathfrak{A}}}
\newcommand{\frakB}{{\mathfrak{B}}}
\newcommand{\frakC}{{\mathfrak{C}}}
\newcommand{\frakD}{{\mathfrak{D}}}
\newcommand{\frakE}{{\mathfrak{E}}}
\newcommand{\frakF}{{\mathfrak{F}}}
\newcommand{\frakG}{{\mathfrak{G}}}
\newcommand{\frakH}{{\mathfrak{H}}}
\newcommand{\frakI}{{\mathfrak{I}}}
\newcommand{\frakJ}{{\mathfrak{J}}}
\newcommand{\frakK}{{\mathfrak{K}}}
\newcommand{\frakL}{{\mathfrak{L}}}
\newcommand{\frakM}{{\mathfrak{M}}}
\newcommand{\frakN}{{\mathfrak{N}}}
\newcommand{\frakO}{{\mathfrak{O}}}
\newcommand{\frakP}{{\mathfrak{P}}}
\newcommand{\frakQ}{{\mathfrak{Q}}}
\newcommand{\frakR}{{\mathfrak{R}}}
\newcommand{\frakS}{{\mathfrak{S}}}
\newcommand{\frakT}{{\mathfrak{T}}}
\newcommand{\frakU}{{\mathfrak{U}}}
\newcommand{\frakV}{{\mathfrak{V}}}
\newcommand{\frakW}{{\mathfrak{W}}}
\newcommand{\frakX}{{\mathfrak{X}}}
\newcommand{\frakY}{{\mathfrak{Y}}}
\newcommand{\frakZ}{{\mathfrak{Z}}}

\newcommand{\fraka}{{\mathfrak{a}}}
\newcommand{\frakb}{{\mathfrak{b}}}
\newcommand{\frakc}{{\mathfrak{c}}}
\newcommand{\frakd}{{\mathfrak{d}}}
\newcommand{\frake}{{\mathfrak{e}}}
\newcommand{\frakf}{{\mathfrak{f}}}
\newcommand{\frakg}{{\mathfrak{g}}}
\newcommand{\frakh}{{\mathfrak{h}}}
\newcommand{\fraki}{{\mathfrak{i}}}
\newcommand{\frakj}{{\mathfrak{j}}}
\newcommand{\frakk}{{\mathfrak{k}}}
\newcommand{\frakl}{{\mathfrak{l}}}
\newcommand{\frakm}{{\mathfrak{m}}}
\newcommand{\frakn}{{\mathfrak{n}}}
\newcommand{\frako}{{\mathfrak{o}}}
\newcommand{\frakp}{{\mathfrak{p}}}
\newcommand{\frakq}{{\mathfrak{q}}}
\newcommand{\frakr}{{\mathfrak{r}}}
\newcommand{\fraks}{{\mathfrak{s}}}
\newcommand{\frakt}{{\mathfrak{t}}}
\newcommand{\fraku}{{\mathfrak{u}}}
\newcommand{\frakv}{{\mathfrak{v}}}
\newcommand{\frakw}{{\mathfrak{w}}}
\newcommand{\frakx}{{\mathfrak{x}}}
\newcommand{\fraky}{{\mathfrak{y}}}
\newcommand{\frakz}{{\mathfrak{z}}}

\newcommand{\setA}{\textsf{A}}
\newcommand{\setB}{\textsf{B}}
\newcommand{\setC}{\textsf{C}}
\newcommand{\setD}{\textsf{D}}
\newcommand{\setE}{\textsf{E}}
\newcommand{\setF}{\textsf{F}}
\newcommand{\setG}{\textsf{G}}
\newcommand{\setH}{\textsf{H}}
\newcommand{\setI}{\textsf{I}}
\newcommand{\setJ}{\textsf{J}}
\newcommand{\setK}{\textsf{K}}
\newcommand{\setL}{\textsf{L}}
\newcommand{\setM}{\textsf{M}}
\newcommand{\setN}{\textsf{N}}
\newcommand{\setO}{\textsf{O}}
\newcommand{\setP}{\textsf{P}}
\newcommand{\setQ}{\textsf{Q}}
\newcommand{\setR}{\textsf{R}}
\newcommand{\setS}{\textsf{S}}
\newcommand{\setT}{\textsf{T}}
\newcommand{\setU}{\textsf{U}}
\newcommand{\setV}{\textsf{V}}
\newcommand{\setW}{\textsf{W}}
\newcommand{\setX}{\textsf{X}}
\newcommand{\setY}{\textsf{Y}}
\newcommand{\setZ}{\textsf{Z}}

\newcommand{\bfa}{\mathbf{a}}
\newcommand{\bfb}{\mathbf{b}}
\newcommand{\bfc}{\mathbf{c}}
\newcommand{\bfd}{\mathbf{d}}
\newcommand{\bfe}{\mathbf{e}}
\newcommand{\bff}{\mathbf{f}}
\newcommand{\bfg}{\mathbf{g}}
\newcommand{\bfh}{\mathbf{h}}
\newcommand{\bfi}{\mathbf{i}}
\newcommand{\bfj}{\mathbf{j}}
\newcommand{\bfk}{\mathbf{k}}
\newcommand{\bfl}{\mathbf{l}}
\newcommand{\bfm}{\mathbf{m}}
\newcommand{\bfn}{\mathbf{n}}
\newcommand{\bfo}{\mathbf{o}}
\newcommand{\bfp}{\mathbf{p}}
\newcommand{\bfq}{\mathbf{q}}
\newcommand{\bfr}{\mathbf{r}}
\newcommand{\bfs}{\mathbf{s}}
\newcommand{\bft}{\mathbf{t}}
\newcommand{\bfu}{\mathbf{u}}
\newcommand{\bfv}{\mathbf{v}}
\newcommand{\bfw}{\mathbf{w}}
\newcommand{\bfx}{\mathbf{x}}
\newcommand{\bfy}{\mathbf{y}}
\newcommand{\bfz}{\mathbf{z}}

\newcommand{\bfalpha}{{\boldsymbol{\alpha}}}
\newcommand{\bfbeta}{{\boldsymbol{\beta}}}
\newcommand{\bfgamma}{{\boldsymbol{\gamma}}}
\newcommand{\bfdelta}{{\boldsymbol{\delta}}}
\newcommand{\bfepsilon}{{\boldsymbol{\epsilon}}}
\newcommand{\bfzeta}{{\boldsymbol{\zeta}}}
\newcommand{\bfeta}{{\boldsymbol{\eta}}}
\newcommand{\bftheta}{{\boldsymbol{\theta}}}
\newcommand{\bfiota}{{\boldsymbol{\iota}}}
\newcommand{\bfkappa}{{\boldsymbol{\kappa}}}
\newcommand{\bflambda}{{\boldsymbol{\lambda}}}
\newcommand{\bfmu}{{\boldsymbol{\mu}}}
\newcommand{\bfnu}{{\boldsymbol{\nu}}}
\newcommand{\bfomicron}{{\boldsymbol{\omicron}}}
\newcommand{\bfpi}{{\boldsymbol{\pi}}}
\newcommand{\bfrho}{{\boldsymbol{\rho}}}
\newcommand{\bfsigma}{{\boldsymbol{\sigma}}}
\newcommand{\bftau}{{\boldsymbol{\tau}}}
\newcommand{\bfupsilon}{{\boldsymbol{\upsilon}}}
\newcommand{\bfphi}{{\boldsymbol{\phi}}}
\newcommand{\bfchi}{{\boldsymbol{\chi}}}
\newcommand{\bfpsi}{{\boldsymbol{\psi}}}
\newcommand{\bfomega}{{\boldsymbol{\omega}}}
\newcommand{\bfxi}{{\boldsymbol{\xi}}}
\newcommand{\bfell}{{\boldsymbol{\ell}}}

\newcommand{\bfA}{\mathbf{A}}
\newcommand{\bfB}{\mathbf{B}}
\newcommand{\bfC}{\mathbf{C}}
\newcommand{\bfD}{\mathbf{D}}
\newcommand{\bfE}{\mathbf{E}}
\newcommand{\bfF}{\mathbf{F}}
\newcommand{\bfG}{\mathbf{G}}
\newcommand{\bfH}{\mathbf{H}}
\newcommand{\bfI}{\mathbf{I}}
\newcommand{\bfJ}{\mathbf{J}}
\newcommand{\bfK}{\mathbf{K}}
\newcommand{\bfL}{\mathbf{L}}
\newcommand{\bfM}{\mathbf{M}}
\newcommand{\bfN}{\mathbf{N}}
\newcommand{\bfO}{\mathbf{O}}
\newcommand{\bfP}{\mathbf{P}}
\newcommand{\bfQ}{\mathbf{Q}}
\newcommand{\bfR}{\mathbf{R}}
\newcommand{\bfS}{\mathbf{S}}
\newcommand{\bfT}{\mathbf{T}}
\newcommand{\bfU}{\mathbf{U}}
\newcommand{\bfV}{\mathbf{V}}
\newcommand{\bfW}{\mathbf{W}}
\newcommand{\bfX}{\mathbf{X}}
\newcommand{\bfY}{\mathbf{Y}}
\newcommand{\bfZ}{\mathbf{Z}}

\newcommand{\bfGamma}{\boldsymbol{\Gamma}}
\newcommand{\bfDelta}{\boldsymbol{\Delta}}
\newcommand{\bfTheta}{\boldsymbol{\Theta}}
\newcommand{\bfLambda}{\boldsymbol{\Lambda}}
\newcommand{\bfPi}{\boldsymbol{\Pi}}
\newcommand{\bfSigma}{\boldsymbol{\Sigma}}
\newcommand{\bfUpsilon}{\boldsymbol{\Upsilon}}
\newcommand{\bfPhi}{\boldsymbol{\Phi}}
\newcommand{\bfPsi}{\boldsymbol{\Psi}}
\newcommand{\bfOmega}{\boldsymbol{\Omega}}

\newcommand{\bbA}{\mathbb{A}}
\newcommand{\bbB}{\mathbb{B}}
\newcommand{\bbC}{\mathbb{C}}
\newcommand{\bbD}{\mathbb{D}}
\newcommand{\bbE}{\mathbb{E}}
\newcommand{\bbF}{\mathbb{F}}
\newcommand{\bbG}{\mathbb{G}}
\newcommand{\bbH}{\mathbb{H}}
\newcommand{\bbI}{\mathbb{I}}
\newcommand{\bbJ}{\mathbb{J}}
\newcommand{\bbK}{\mathbb{K}}
\newcommand{\bbL}{\mathbb{L}}
\newcommand{\bbM}{\mathbb{M}}
\newcommand{\bbN}{\mathbb{N}}
\newcommand{\bbO}{\mathbb{O}}
\newcommand{\bbP}{\mathbb{P}}
\newcommand{\bbQ}{\mathbb{Q}}
\newcommand{\bbR}{\mathbb{R}}
\newcommand{\bbS}{\mathbb{S}}
\newcommand{\bbT}{\mathbb{T}}
\newcommand{\bbU}{\mathbb{U}}
\newcommand{\bbV}{\mathbb{V}}
\newcommand{\bbW}{\mathbb{W}}
\newcommand{\bbX}{\mathbb{X}}
\newcommand{\bbY}{\mathbb{Y}}
\newcommand{\bbZ}{\mathbb{Z}}
\newcommand{\prl}[1]{\left(#1\right)}
\newcommand{\brl}[1]{\left[#1\right]}
\newcommand{\crl}[1]{\left\{#1\right\}}


\title{Curriculum Imitation Learning of Distributed Multi-Robot Policies\\
}

\author{
\IEEEauthorblockN{
Jes\'{u}s Roche\IEEEauthorrefmark{1},
Eduardo Sebasti\'{a}n\IEEEauthorrefmark{2}, 
Eduardo Montijano\IEEEauthorrefmark{1}
}
\IEEEauthorblockA{
\IEEEauthorrefmark{1}Departamento de Inform\'atica e Ingenier\'ia de Sistemas, Universidad de Zaragoza, Zaragoza, Spain\\
Emails: \{j.roche, emonti\}@unizar.es
}
\IEEEauthorblockA{
\IEEEauthorrefmark{2}Department of Computer Science and Technology, University of Cambridge, Cambridge, United Kingdom\\
Email: es2121@cam.ac.uk
}
\thanks{This work has been supported by the ONR Global grant N62909-24-1-2081, Spanish project PID2024-159284NB-I00 funded by MCIN/AEI/10.13039/501100011033, by ERDF A way of making Europe and by the European Union NextGenerationEU/PRTR, and DGA T45-23R.}
}

\maketitle
\copyrightnotice

\begin{abstract}
Learning control policies for multi-robot systems (MRS) remains a major challenge due to long-term coordination and the difficulty of obtaining realistic training data. 
In this work, we address both limitations within an imitation learning framework. First, we shift the typical role of Curriculum Learning in MRS, from scalability with the number of robots, to focus on improving long-term coordination. We propose a curriculum strategy that gradually increases the length of expert trajectories during training, stabilizing learning and enhancing the accuracy of long-term behaviors. Second, we introduce a method to approximate the egocentric perception of each robot using only third-person global state demonstrations. Our approach transforms idealized trajectories into locally available observations by filtering neighbors, converting reference frames, and simulating onboard sensor variability. Both contributions are integrated into a physics-informed technique to produce scalable, distributed policies from observations. We conduct experiments across two tasks with varying team sizes and noise levels. Results show that our curriculum improves long-term accuracy, while our perceptual estimation method yields policies that are robust to realistic  uncertainty. Together, these strategies enable the learning of robust, distributed controllers from global demonstrations, even in the absence of expert actions or onboard measurements.

\end{abstract}


\section{Introduction}\label{sec:intro}
Multi-robot systems (MRS) outperform their single-robot counterparts, increasing fault tolerance, and efficiency. 
Hand-crafting each control policy to coordinate the robots is often complex and requires expert knowledge. For this reason, data-driven methods are often used as an effective path to learn control policies.
Recent advances in graph-based neural policies~\cite{sebastian2023lemurs, sebastian2024physics,  Khan2021, Hu2025} are a good example of how machine learning can offer a data-driven path to scalability and distribution. 
Yet, current multi-robot learning methods still present major challenges. \textit{Long-term coordination} is hindered by the potential conflicts between individual robot objectives, specially in the absence of a mathematical description of the task such as in imitation learning \cite{osa2018algorithmic} and multi-agent reinforcement learning \cite{chen2021variational}. 
Existing solutions limit their supervision to a few samples at a time, due to the high dimensionality of
the search space and stability problems.
Curriculum Learning \cite{wang2021survey} emerges as a promising approach to address this challenge: the acquisition of complex behaviors is facilitated by gradually increasing the complexity of tasks during training \cite{bengio2009curriculum, bhattacharyya2018multi, long2020evolutionary}. 

Secondly, obtaining the necessary data to supervise the training in imitation learning is very complex and expensive, as it requires knowledge of the actions and observations taken by each and all agents simultaneously.
Learning from observations~\cite{Boborzi2023} is a paradigm inside imitation learning that relies exclusively on external observations (e.g., videos of expert behavior) as learning data. 
Global state trajectories like third-person video of the full scene are far more available than first-person sensor data, such as LiDAR measurements or egocentric videos of each individual agent.
While this paradigm enables imitation from human or animal demonstrations, the control actions and local perception of the expert are not available to learn the policy, making it difficult to implement them in robots with onboard imperfect sensing.

To overcome these challenges, in this work we introduce a Curriculum Learning approach to gradually increase the length of expert trajectories during training, stabilizing the learning process and leading to accurate long-term behavior reproduction.
Integrating our long-term coordination strategy on LEMURS~\cite{sebastian2023lemurs} allows us to shift the perspective away from scalability, since it is a strategy used to learn scalable distributed policies.
Additionally, to bridge the gap caused by third-person observations, we propose a transformation to simulate noisy onboard measurements from the (almost) perfect global observation.
This approach penalizes overly confident or reactive behaviors and yields policies that are more robust to perceptual uncertainty.

We demonstrate the effectiveness of our approach through extensive experiments that involve two task settings and six configurations of varying noise levels and robot sizes.  
We show that our curriculum design leads to improved long-term accuracy of the learned distributed multi-robot policies compared to their non-curriculum counterparts. We also show that policies learned considering noisy observations are more robust and less likely to drift from the expert behavior.

\section{Related Work}\label{sec:related}

In the context of multi-robot systems, Curriculum Learning is mainly used to address scalability on the number of robots
\cite{bhattacharyya2018multi, long2020evolutionary, yang2020cm3}. 
An interesting second option is to train a single robot to solve the task, and then use Curriculum Learning to move to the multi-robot setting, where the robots learn how to cooperate \cite{yang2020cm3}. Another way of applying Curriculum Learning to multi-robot settings is for increasing the environmental complexity \cite{narvekar2020curriculum}. For instance, in pursuit-evasion games \cite{de2021decentralized, qu2023pursuit}, the robots are firstly presented with easy-to-capture targets, gradually improving the escaping capabilities of the evaders and the obstacle density of the scenario. More generally, curriculum can be used to gradually select goals with increasing configuration complexity in manipulation or navigation tasks \cite{florensa2017reverse}. In path planning problems, traditional planning methods can be used to decompose a path planning task in sub-goals and design a curriculum that schedules the sub-goals from local easy-to-plan tasks to the original planning problem \cite{ao2021co}. Differently, we focus our attention on capturing long-term behaviors from expert demonstrations by designing a curriculum on the length of the expert trajectories. This focus is crucial for long-term coordination, since short demonstrations capture only local interactions.

Closely related to our solution, \cite{liu2021curriculum,feng2022curriculum,blessing2024information} propose an imitation learning framework where the control policy and the curriculum are learned simultaneously. 
Unlike these works, which define curricula over samples, goals, or tasks in non-robotic or single-agent settings, ours is the first work that proposes Curriculum Learning to enhance imitation learning for distributed multi-robot policies, by leveraging the temporal horizon during training.
The aforementioned works lie in the category of adaptive/automatic Curriculum Learning \cite{chen2021variational, kang2023learning}, where a heuristic (typically based on neural networks) learns how to tune the parameters of different base automatic curriculum modules that schedule, e.g., the initial robot configuration, the sub-goals, or the reward in reinforcement learning settings. In this work, we opt for a predefined heuristic in order to restrict the training complexity to the distributed control policy and guarantee that the learned policies remain to be scalable with the number of robots.


\section{Preliminaries}\label{sec:prosta}
\subsection{Learning distributed multi-robot policies from observations}
\label{sec:imitation_learning}
We consider a group of $n$ robots, denoted as \mbox{$\mathcal{V} = \{1, \hdots, n\}$}.  The state of robot $i$ at time $t \geq 0$ is denoted as $\mathbf{x}_i(t) \in \mathbb{R}^{n_x}$. The stack of all robots compose the state of the system, $\mathbf{x}(t) = [\mathbf{x}_1(t), \hdots, \mathbf{x}_n(t)]$. Similarly, joining the input control $\mathbf{u}_i(t) \in \mathbb{R}^{n_u}$ of every robot results in the system control input,  $\mathbf{u}(t) = [\mathbf{u}_1(t), \hdots, \mathbf{u}_n(t)]$.
We assume known system dynamics, $\dot{\mathbf{x}}(t) = f(\mathbf{x}(t), \mathbf{u}(t)),$
defined by each robot dynamics $\dot{\mathbf{x}}_i(t) = f_i(\mathbf{x}_i(t), \mathbf{u}_i(t))$.
Robots interact in a distributed manner. We model the topology of interactions using an adjacency matrix \mbox{$\mathbf{A}(t) \in \{0,1\}^{n \times n}$} such that $\mathbf{A}_{ij}(t) = 1$ if robots $i$ and $j$ exchange information at time $t$, and  $\mathbf{A}_{ij}(t) = 0$ otherwise. We define a set of neighbors for robot $i$ at time $t$ as $\mathcal{N}_i(t) = \{j \in \mathcal{V} \hspace{0.5em} | \hspace{0.5em} \mathbf{A}_{ij}(t) = 1 \}$. Note that a robot is always considered to be its own neighbor.

Each robot $i$ executes an unknown distributed control policy $\pi_\bftheta$ that only depends on the local observation $\mathbf{y}_{i}(t)$, which captures a perceptually filtered and potentially noisy view of the neighboring states $\mathbf{x}_{\mathcal{N}_i}$.
\begin{equation}\label{eq:defPoliticaControl}
        \mathbf{u}_i(t) = \pi_\bftheta(\mathbf{y}_{i}(t)), 
\end{equation}
where $\bftheta$ represents the control policy parameters.

In the context of imitation learning, the goal is to learn the parameters $\bftheta$ of a policy $\pi_\bftheta$ from task demonstrations. The learned policy should drive the system's state evolution to meet the objectives shown in the demonstrations, while respecting the system dynamics $\dot{\mathbf{x}}(t)$. 

Learning from observations is a particular case of imitation learning where control actions $\mathbf{u}_i(t)$ and local observations $\mathbf{y}_{i}(t)$ of the expert are not available, and we rely solely on their global  state of the group for the learning process. This is our case, since we only assume availability of demonstrated trajectories. 

For an initial state $\mathbf{x}(0) = \overline{\mathbf{x}}(0)$, its demonstrated trajectory is $\overline{\mathbf{x}}_{0:K} = [\overline{\mathbf{x}}(0), \hdots, \overline{\mathbf{x}}(kT), \hdots, \overline{\mathbf{x}}(KT)]$, where $K$ is the number of samples and $T$ the sampling period. Given $L$ demonstrated trajectories, indexed by $l$, they are grouped in the training dataset $\overline{\mathcal{D}}_K = \{\overline{\mathbf{x}}_{0:K}^l\}_{l=0}^L$. Similarly, the trajectories predicted by the imitation learning model are denoted as $\mathbf{x}_{0:K} = [\mathbf{x}(0), \hdots, \mathbf{x}(kT), \hdots, \mathbf{x}(KT)]$, defining the set  $\mathcal{D}_K = \{\mathbf{x}_{0:K}^l\}_{l=0}^L$. We aim to learn a control policy that minimizes the distance between the demonstrated and predicted trajectories,
\begin{equation}\label{eq:loss_function}
    \mathcal{L}({\mathcal{D}_K},\overline{\mathcal{D}}_K) = \frac{1}{KL}\sum_{l=0}^L\sum_{k=0}^K
    ||\mathbf{x}^l(kT)-\overline{\mathbf{x}}^l(kT)||^2_2.
\end{equation}
Taking this into account, the formal objective of imitation learning in this context is to solve the following optimization problem,
\begin{subequations}\label{eq:prob_def}
\begin{alignat}{2}
\min_{\bftheta}
& \hbox{ }       \mathcal{L}({\mathcal{D}_K},\overline{\mathcal{D}}_K) \label{eq:prob_def_cost}
\\
\text{s.t.}
& \hbox{ }\dot{\mathbf{x}}_{i}^l(t) = f_i(\mathbf{x}_i^l(t), \mathbf{u}_i^l(t)), \hbox{ } \bfx_i^l(0) = \overline{\bfx}_i^l(0), \hbox{ } \forall i,l,\label{eq:prob_def_constraint2}
\\
& \hbox{ } {\mathbf{u}}_i^l(t) = \bfpi_{\bftheta}(h_i(\mathbf{x}^l(t))), \hbox{ } \forall i,l,t. \label{eq:prob_def_constraint3}
\end{alignat}
\end{subequations}
where $h_i(\mathbf{x})$ is the function that robot $i$ uses to estimate $\mathbf{y}_i$. Additional details about the formulation of multi-robot imitation learning used in this paper can be found in~\cite{sebastian2023lemurs}.
Unfortunately, if the value of $K$ in the dataset is large, the high dimensionality of the problem and the cumulative nature of the loss between the sequential states of the predicted and demonstrated trajectories leads to exploding gradients. The consequence is either 
convergence to wrong policy parameters that display poor long-term behaviors, or even an unstable training. In this paper, we show how to design a Curriculum Learning approach to overcome this problem.
Similarly, the lack of direct supervision on the onboard observations, $\mathbf{y}_i^l$, requires to build a suitable approximation from the predicted states $h_i(\mathbf{x}^l(t))$ .

\subsection{Curriculum Learning}
\label{sec:curriculum_prelim}

Curriculum Learning (CL) is a learning paradigm where an agent is trained iteratively following a curriculum to ease learning in difficult problems, such as~\eqref{eq:prob_def}. A curriculum is a sequence of training criteria over $E$ training steps, \mbox{$\mathcal{Q}=\{Q_1,\ldots,Q_E\}$, where $Q_e$, for $e=1,\ldots,E,$} denotes the training criterion at training step $e$. The training criterion encompasses the training parameters and features that characterize the complexity of the task, e.g., the features of the scenarios that compose the training set, the optimization goals included in the loss function, or the set of model parameters being optimized~\cite{wang2021survey}.

Therefore, the goal of this paper is to define a curriculum $\mathcal{Q}$ to overcome some of the difficulties of optimizing $\bftheta$ in~\eqref{eq:prob_def}. In particular, and different from existing works that primarily focus on designing curricula to address the complexity associated with the number of robots, we propose to apply Curriculum Learning in the length of the expert trajectories $K$ to enable accurate long-term behaviors.


\section{Methodology}\label{sec:solution}
\subsection{Curriculum Learning for accurate long-term behaviors} \label{sec:cl_solution}
This section presents the curriculum strategy to improve the training performance in imitation learning for multi-robot problems.
Following~\cite{wang2021survey}, we identify three key elements that we need to specify for each $Q_e\in\mathcal{Q}:$ (i) a Difficulty Measurer to sort the dataset in elements of increasing difficulty for the learning algorithm, (ii) a Training Scheduler to describe when we should increase the difficulty and for how many training steps and (iii) a loss function that can adapt to the difficulty of the dataset at each training stage.
In this paper we propose a manually predefined curriculum design and describe each of the three elements in detail in the following subsections.  

\subsubsection{Difficult Measurer}\label{sec:diffic_measurer}

The Difficulty Measurer is responsible for sorting the available data according to its complexity. 
It uses the length of the trajectories 
to sort the dataset based on the difficulty.
We denote by $K_e$ the maximum trajectory length considered for the training criterion $Q_e$.

Since it might be difficult to have example trajectories of every required length, $K_e$, we propose a partitioning method of the existing trajectories to obtain such data at each training stage. Besides, with this approach we are also able to 
fully exploit the information provided by the expert trajectories. In this sense, we assume that the task and the trajectories respect the Markov property, i.e., the evolution of the trajectories over time does not depend on its history. In our context, this implies that transition from $\overline{\mathbf{x}}^l(k)$ to $\overline{\mathbf{x}}^l(k+1)$ only depends on the configuration of the system at sampled instant $k$.

Given a trajectory of the dataset $\overline{\mathcal{D}}_K$,  \mbox{$\overline{\mathbf{x}}_{0:K}^l = [\overline{\mathbf{x}}^l(0), \hdots, \overline{\mathbf{x}}^l(KT)]$}, we generate a sub-trajectory by randomly picking an initial time step, $k_0$, in the interval $[0,K-K_e].$ Then, we obtain a demonstrated trajectory of length $K_e$ by considering the following $K_e$ states in the data, 
\begin{equation}\label{eq:diff_me}
    \overline{\mathbf{x}}_{0:K_e}^l = [\overline{\mathbf{x}}^l(k_0T), \hdots,  \overline{\mathbf{x}}^l((k_0+K_e)T)].
\end{equation}
Following this procedure multiple times we generate the training dataset of difficulty $K_e,$ $\overline{\mathcal{D}}_{K_e}$.

\subsubsection{Training Scheduler}\label{sec:train_schedule}

The Training Scheduler is in charge of defining the training step in which the curriculum transitions from training criterion $Q_e$ to training criterion $Q_{e+1}$.
We denote by $N_e$ the number of training steps executed with criterion $Q_e$.
Our proposed design uses a discrete \emph{Baby Step} approach to model the evolution of $N_e$ and the level of difficulty, specified by $K_e$.
In this algorithm, the increments are equal for every training step, identified by the parameters $c_K$ and $c_N$ respectively.
This simple procedure is useful to capture settings where the complexity on the length of the trajectories associated to the loss function and policy parameters increases linearly, which is a reasonable assumption for the case of Eq.~\eqref{eq:defPoliticaControl}. The reasoning behind the schedule on $N_e$ follows a closely related motivation, where the proposed curriculum switches between training criteria every $c_N$ steps. 
We deliberately adopt this simple schedule to isolate the effect of trajectory length on the learning process, avoiding additional complexity from adaptive mechanisms. A more elaborated Training Scheduler is left for future work.

\subsubsection{Training Loss}\label{sec:train_loss}

We update the general loss function in Eq. \eqref{eq:loss_function} to reflect the time-horizon's dynamic behavior along the training. The proposed dynamic training loss normalizes the error by the time-horizon $K_e$, 
\begin{equation}
\label{eq:dynamic_loss}
    \mathcal{L}_e(\mathcal{D}_{K_e}, \overline{\mathcal{D}}_{K_e}) = 
    \frac{1}{K_eL} \sum_{l=0}^L\sum_{k=0}^{K_e} || \mathbf{x}^l(kT) - \overline{\mathbf{x}}^l(kT) ||_2^2,
\end{equation}
to make sure that gradients are informative enough in all training stages, but especially during the first training steps, when the value of $K_e$ is much smaller than $K$.

\subsubsection{CL Algorithm}
\begin{figure}
    \centering
    \includegraphics[width=0.7\linewidth]{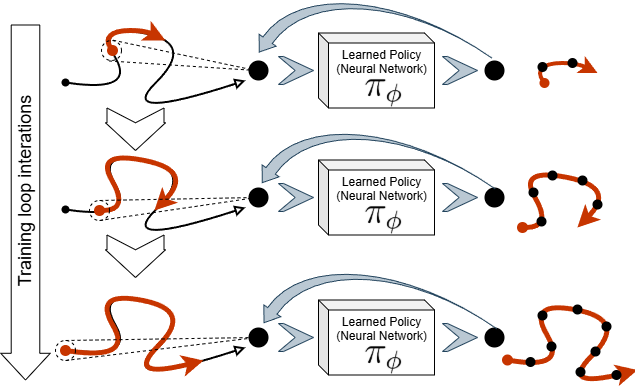}
    \caption{\small CL for long-term behaviors. We start predicting short trajectories and gradually increase the duration.}
    \label{fig:CL_scheme}
\end{figure}

In summary, $Q_e$ is identified by three parameters: (i) $K_e$, which describes the complexity of the trajectories used to train at that training stage $e$; (ii) the training interval, $N_e$, measured in training steps; (iii) and the training loss, $\mathcal{L}_e$. 
We remark that, as with other Curriculum approaches, 
ours can be seamlessly integrated in any multi-robot imitation learning setting, independent on the specific policy parameterization. 
Figure~\ref{fig:CL_scheme} shows graphically the idea of the curriculum algorithm.

\subsection{Estimating egocentric perception from global state demonstrations} \label{sec:noise_solution}

Real-world multi-robot deployments rely on distributed decision-making based on each robot's local and often noisy perception of its neighborhood. However, in many imitation learning settings, such as those using video demonstrations of natural swarms, only clean, third-person state trajectories are available. These global demonstrations do not reflect what individual robots would perceive from their own limited and noisy viewpoints. Training policies directly from this data can lead to poor generalization and induce lack of robustness when robots are deployed in real environments.

In this section, we describe our approach to estimate each robot's local observation from third-person state demonstrations. This approximation serves as a substitute for real egocentric measurements during training, enabling us to train perception-driven distributed policies from global trajectories. The process consists of three key steps: First, selecting relevant neighboring states to simulate limited perception. Second, transforming those states into the robot's local reference frame to emulate egocentric sensing. Finally, we model the variability caused by sensor imperfections by injecting mild stochastic perturbations.

As shown in Eq.~\eqref{eq:defPoliticaControl}, the distributed policy $\pi_\bftheta$ takes as input a robot's local observation $\mathbf{y}_i(t)$ and outputs the corresponding control action $\mathbf{u}_i(t)$. However, during training we only have access to the global system state $\mathbf{x}(t)$.
To approximate $\mathbf{y}_i(t)$ from global data, we first extract the state of neighboring robots, $\mathbf{x}_{\mathcal{N}_i}(t)$. Then, we express these relative to the position of the robot, emulating the information that would be locally accessible through perception or short-range communication. Finally, we simulate sensor uncertainty by perturbing the result with Gaussian noise.
Formally, we define the estimated local observation as
\begin{equation}
    h_i(\mathbf{x}(t)) = \mathbf{x}_{\mathcal{N}_i}(t) - \mathbf{x}_i(t) + \boldsymbol{\eta}_i(t), \quad
    \boldsymbol{\eta}_i(t) \sim \mathcal{N}(0, \sigma^2 \mathbf{I}),
\end{equation}
where $\boldsymbol{\eta}_i(t) \in \mathbb{R}^{d_i}$ is a zero-mean Gaussian noise vector with standard deviation $\sigma$, $\mathbf{I}$ is the identity matrix and $d_i$ is the dimension of the input observation for robot $i$. This formulation is intentionally simplistic for efficiency and clarity. In practice, onboard perception modules like YOLO \cite{redmon2016you} similarly provide relative neighbor estimates subject to stochastic noise, which motivates our assumption that egocentric observations can be derived from global states. Future work could integrate complex perceptual effects such as occlusions.

The control action is thus evaluated on the estimated observation for all predictions involved in the training  process of optimization:
\begin{equation}
    \mathbf{u}_i(t) = \pi_\bftheta\big(  h_i(\mathbf{x}(t))\big),
\end{equation}
and fed in~\eqref{eq:prob_def_constraint2} to obtain the predicted global states used in the learning process.


\section{Experiments}\label{sec:simulations}
This section evaluates the proposed technique over various problem settings. We integrate our proposed Curriculum Learning algorithm and noise injection module in LEMURS \cite{sebastian2023lemurs}, a physics-informed policy parameterization specially tailored for distributed multi-robot problems\footnote{The code to reproduce the results of the paper can be found in the following repository: \url{https://github.com/jesusico2001/Curriculum-Imitation-Learning-MRS}}. 

\subsection{Experimental setup}
We consider a 2-dimensional space for all scenarios, defining the state of a robot as $\mathbf{x}_i(t) = [\mathbf{p}_i(t), \mathbf{v}_i(t)] \in \mathbb{R}^{4}$, which contains the robot's position $\mathbf{p}_i(t) \in \mathbb{R}^{2}$ and velocity $\mathbf{v}_i(t) \in \mathbb{R}^{2}$.
We rely on VMAS\cite{Bettini2024}, a vectorized multi-robot 2D simulator, to provide environments with meaningful tasks that require agent coordination. In order to evaluate the proposed curriculum and noise injection strategies, the two selected tasks are: 
\begin{enumerate}
    \item \textit{Navigation}: initially, robots and goals are randomly arranged in a $5$m$\times5$m area, ensuring that there is no overlapping. Then, each robot is required to navigate to its goal while avoiding collisions.
    \item \textit{Passage}: the task to accomplish is similar to \textit{navigation}, but the map is split horizontally by a wall that robots can only traverse through a randomly placed passage. Robots are initially grouped in the bottom half, and goals in the top half, such that they are forced to navigate through the passage in a cluttered space.
\end{enumerate}
Examples of these tasks are given in Figure~\ref{fig:qualitative} and further details about them can be found in~\cite{Bettini2024}.

For each task, we generate a training dataset composed by $L=5000$ trajectories of length $K=200$ for \textit{navigation} and $K=300$ for \textit{passage}, to ensure that experts have time to finish the task, the sampling period is $T=0.05$s. The trajectories for the \textit{navigation} task are generated using a centralized RL policy trained using BenchMARL \cite{benchmarl} with the default hyperparameters. For \textit{passage}, we designed an analytical controller that combines potential fields for reactive collision avoidance, with a fixed path planner that sets intermediate goals before and after the passage to align the robot velocities. 
We train models using demonstrations of $n=6$ and $n=12$ robots. With levels of noise $\sigma= 0$,  $\sigma= 0.1$ and $\sigma= 0.25$.

The curriculum proposed 
is parameterized with $c_K = 1$, and $c_N = 150$ epochs.
A non-curriculum training with a fixed horizon of $K=5$ serves as a baseline to measure improvements.
All the models are trained using the Adam optimizer with a learning rate of $0.005$ for $E = 5000$ training steps.

\subsection{Metrics}
We consider four different metrics for evaluation:
\begin{enumerate}
    \item The \textbf{square euclidean loss} $\mathcal{L}$ defined in Eq. \eqref{eq:loss_function} is the training objective and accounts for errors in position and velocity; outliers have a strong impact due to its quadratic nature.
    
    \item The \textbf{mean position error} $\mathcal{E}_\text{pos}$ is similar to the loss, but it does not regard velocities and outliers impact linearly. It is computed as follows:
    \begin{equation}\label{eq:mean_pos_error}
        \mathcal{E}_{\text{pos}} = \frac{1}{KL} \sum_{l=0}^{L} \sum_{k=0}^{K}
        \frac{1}{n} \sum_{i=1}^{n} \left\| \mathbf{p}_i^l(kT) - \overline{\mathbf{p}}_i^l(kT) \right\|_2,
    \end{equation}
    where $\mathbf{p}_i(kT)$ and $\Bar{\mathbf{p}}_i(kT)$ denote the predicted and expert positions of robot $i$ at time $kT$.

    \item The \textbf{Fr\'{e}chet distance} $\mathcal{F}$ between predicted and demonstrated trajectories, to account for both spatial proximity and temporal alignment.  
    We compute it using Similarity Measures\cite{area_trajectories}.

    \item The \textbf{number of tasks completed} $n_\text{comp}$, or robots that are close to their goal by the end of each episode, within a tolerance error of $0.25$ distance units (which corresponds to the maximum measurement noise). This metric reflects the collective reliability of the policy, its robustness and the presence of failure cases or outliers.

\end{enumerate}
The results reported in the paper are the average of the metrics over $200$ randomly selected unseen test trajectories.

\subsection{Results}
\begin{table}[!ht]
    \centering
    \begin{tabular}{|c|c|c|c|c|c|c|c|}
\hline
Task& $n$ & $\sigma$ & CL & $\mathcal{L}$ & $\mathcal{E}_\text{pos}$ & $\mathcal{F}$ & $n_\text{comp}$ \\
\hline
\multirow{12}{*}{\rotatebox[origin=c]{90}{\textit{navigation}}} & \multirow{6}{*}{$6$} & \multirow{2}{*}{$0$} & No & $0.071$ & $0.373$ & $0.391$ & $1.005$ \\ 
                             &  &   & Yes & $\mathbf{0.055}$ & $\mathbf{0.314}$ & $\mathbf{0.366}$ & $\mathbf{1.485}$\\
\cline{3-8}
           &  & \multirow{2}{*}{$0.1$} & No & $0.067$ & $0.388$ & $0.402$ & $2.925$\\
           &  &   & Yes & $\mathbf{0.056}$ & $\mathbf{0.328}$ & $\mathbf{0.378}$ & $\mathbf{3.825}$\\
\cline{3-8}
                             &  & \multirow{2}{*}{$0.25$} & No & $0.072$ & $0.383$ & $0.397$ & $5.405$\\
                             &  &   & Yes & $\mathbf{0.059}$ & $\mathbf{0.333}$ & $\mathbf{0.385}$ & $\mathbf{5.740}$\\
\cline{2-8}
                             & \multirow{6}{*}{$12$}  & \multirow{2}{*}{$0$} & No & $0.054$ & $0.343$ & $0.388$ & $2.020$\\
                             &   &   & Yes & $\mathbf{0.041}$ & $\mathbf{0.280}$ & $\mathbf{0.353}$ & $\mathbf{3.275}$\\
\cline{3-8}
                             &   & \multirow{2}{*}{$0.1$} & No & $0.049$ & $0.328$ & $0.383$ & $6.605$\\
                            &   &   & Yes & $\mathbf{0.040}$ & $\mathbf{0.276}$ & $\mathbf{0.345}$ & $\mathbf{8.375}$\\
\cline{3-8}
                             &   & \multirow{2}{*}{$0.25$} & No & $0.050$ & $0.328$ & $0.383$ & $10.920$\\
                             &   &   & Yes & $\mathbf{0.041}$ & $\mathbf{0.279}$ & $\mathbf{0.347}$ & $\mathbf{11.695}$\\
\hline
\hline
\multirow{12}{*}{\rotatebox[origin=c]{90}{\textit{passage}}}  & \multirow{6}{*}{$6$} & \multirow{2}{*}{$0$} & No & $0.225$ & $0.666$ & $1.050$ & $0.345$\\
  &   &   & Yes & $\mathbf{0.101}$ & $\mathbf{0.441}$ & $\mathbf{0.621}$ & $\mathbf{0.865}$\\
\cline{3-8}
  &   & \multirow{2}{*}{$0.1$} & No & $0.247$ & $0.688$ & $1.060$ & $1.025$\\
  &   &   & Yes & $\mathbf{0.191}$ & $\mathbf{0.564}$ & $\mathbf{0.912}$ & $\mathbf{2.365}$\\
\cline{3-8}
  &   & \multirow{2}{*}{$0.25$} & No & $0.230$ & $0.650$ & $1.018$ & $2.375$\\
  &   &   & Yes & $\mathbf{0.223}$ & $\mathbf{0.613}$ & $\mathbf{0.955}$ & $\mathbf{3.315}$\\
\cline{2-8}
  & \multirow{6}{*}{$12$} & \multirow{2}{*}{$0$} & No & $0.244$ & $0.731$ & $1.162$ & $0.155$\\
  &   &   & Yes & $\mathbf{0.187}$ & $\mathbf{0.623}$ & $\mathbf{0.920}$ & $\mathbf{0.430}$\\
\cline{3-8}
  &   & \multirow{2}{*}{$0.1$} & No & $0.305$ & $0.834$ & $1.372$ & $0.280$\\
  &   &   & Yes & $\mathbf{0.126}$ & $\mathbf{0.493}$ & $\mathbf{0.703}$ & $\mathbf{2.745}$\\
\cline{3-8}
  &   & \multirow{2}{*}{$0.25$} & No & $0.197$ & $0.662$ & $1.024$ & $1.395$\\
 &   &   & Yes & $\mathbf{0.150}$ & $\mathbf{0.542}$ & $\mathbf{0.793}$ & $\mathbf{4.185}$\\
\hline
    \end{tabular}
    \caption{\small Results for the two tasks and six configurations of group size $n$ and noise level $\sigma$, with and without Curriculum Learning (best in bold).}
    \label{tab:table_results}
\end{table}

\begin{figure}
    \centering
    \includegraphics[trim={90, 40 70, 60},clip,width=0.49\linewidth]{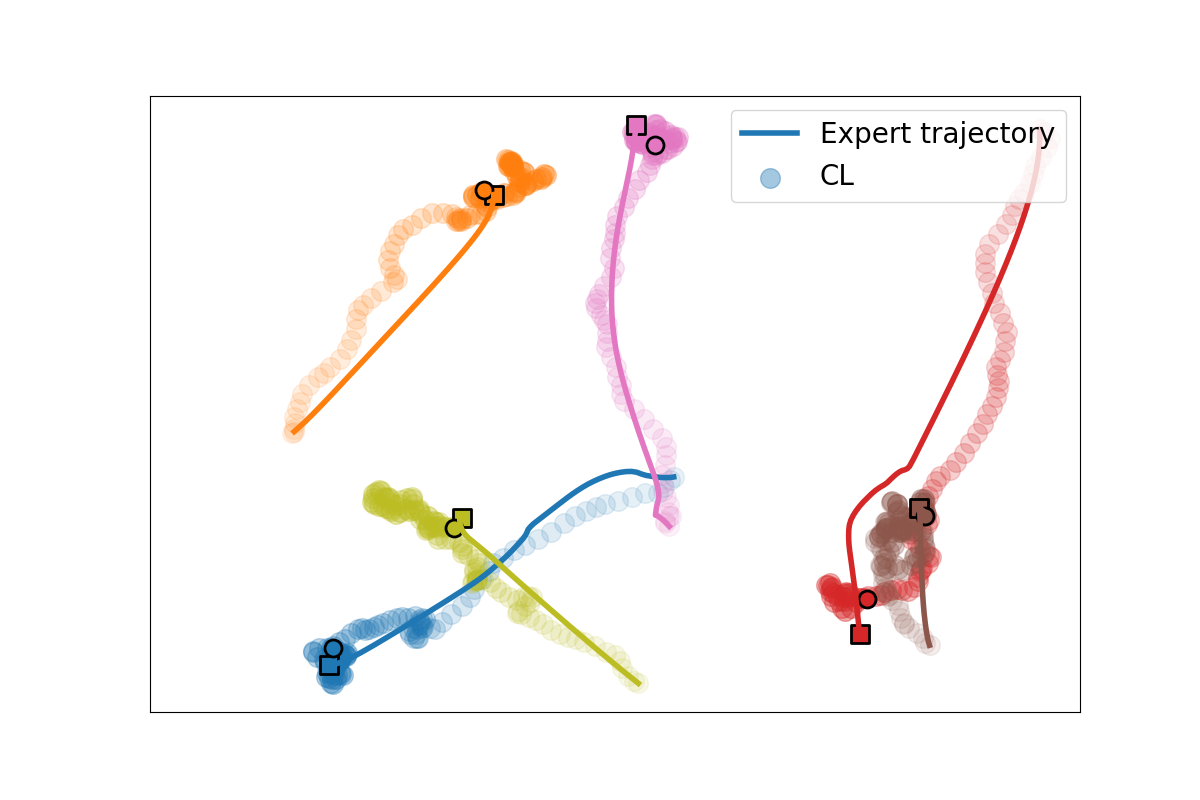}
    \includegraphics[trim={90, 40 70, 60},clip,width=0.49\linewidth]{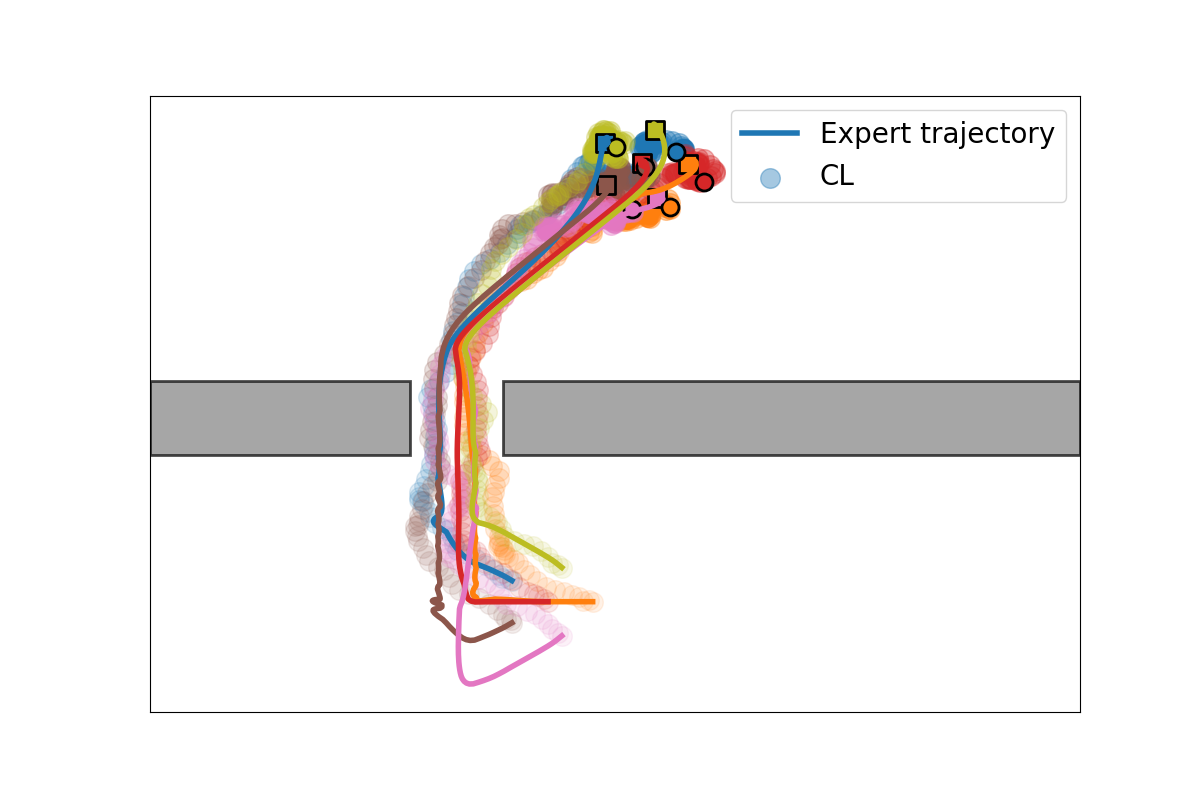}
    \\
    \includegraphics[trim={90, 40 70, 60},clip,width=0.49\linewidth]{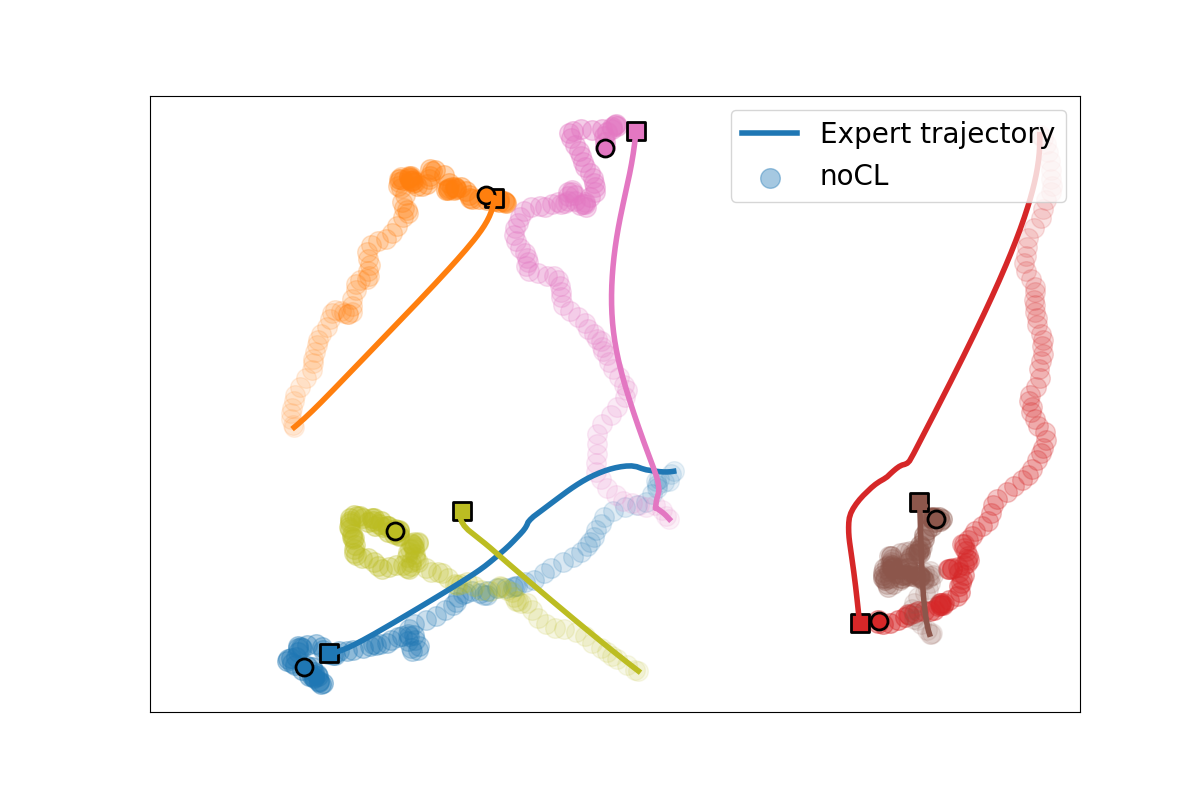}
    \includegraphics[trim={90, 40 70, 60},clip,width=0.49\linewidth]{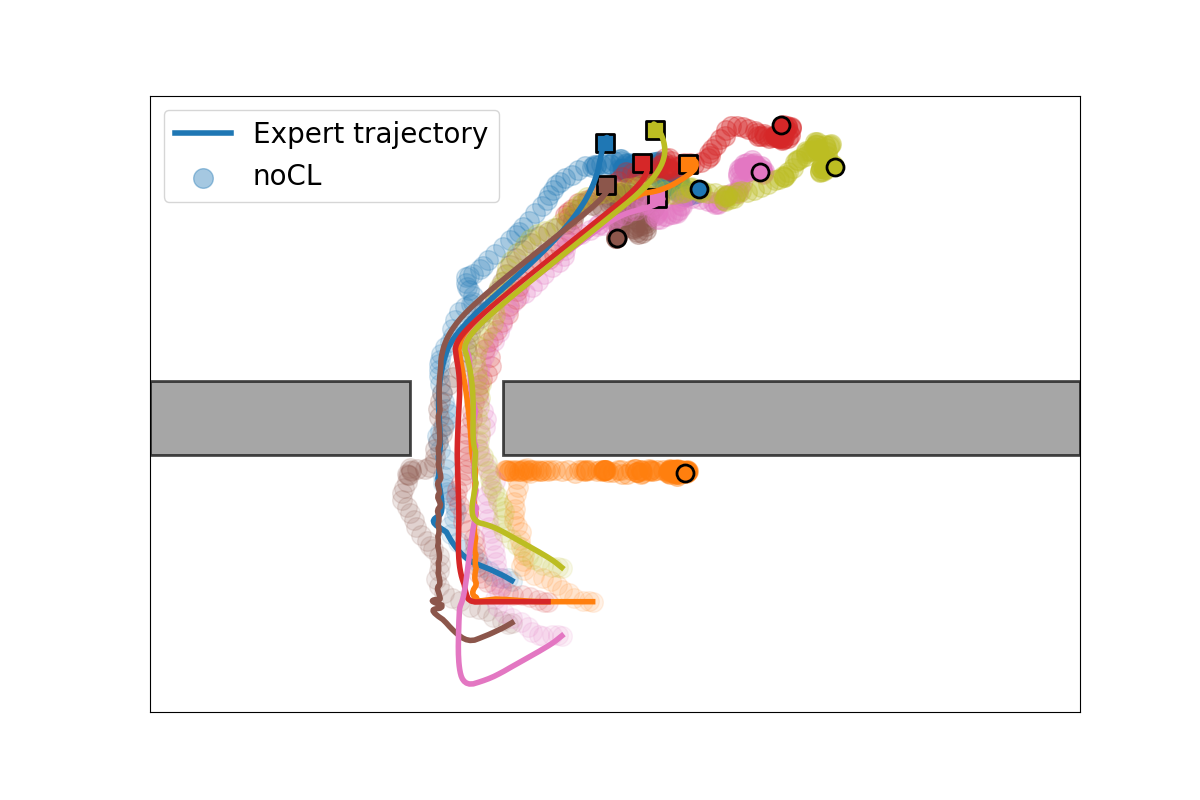}

    \caption{\small Qualitative visualization of the predicted trajectories  in \textit{navigation} (left) and \textit{passage} (right) for $n=6$ and $\sigma=0.25$. The solid lines and the square marks correspond to the demonstrated trajectories and their final positions, respectively. 
    The trajectories that correspond to the policy trained with curriculum (top) and without CL (bottom) are visualized using circle marks  and their final positions have a black border. Each color represents a robot. }
    \label{fig:qualitative}
\end{figure}

Table \ref{tab:table_results} shows the evaluation metrics for the \textit{navigation} and \textit{passage} tasks, each evaluated under the same noise level $\sigma$ used for training. Across all metrics, CL consistently outperforms the non-curriculum baseline, this is also reflected on the robot trajectories show in Fig. \ref{fig:qualitative}.
\begin{figure}
\includegraphics[trim={0, 0 70, 60},clip,width=0.49\linewidth]{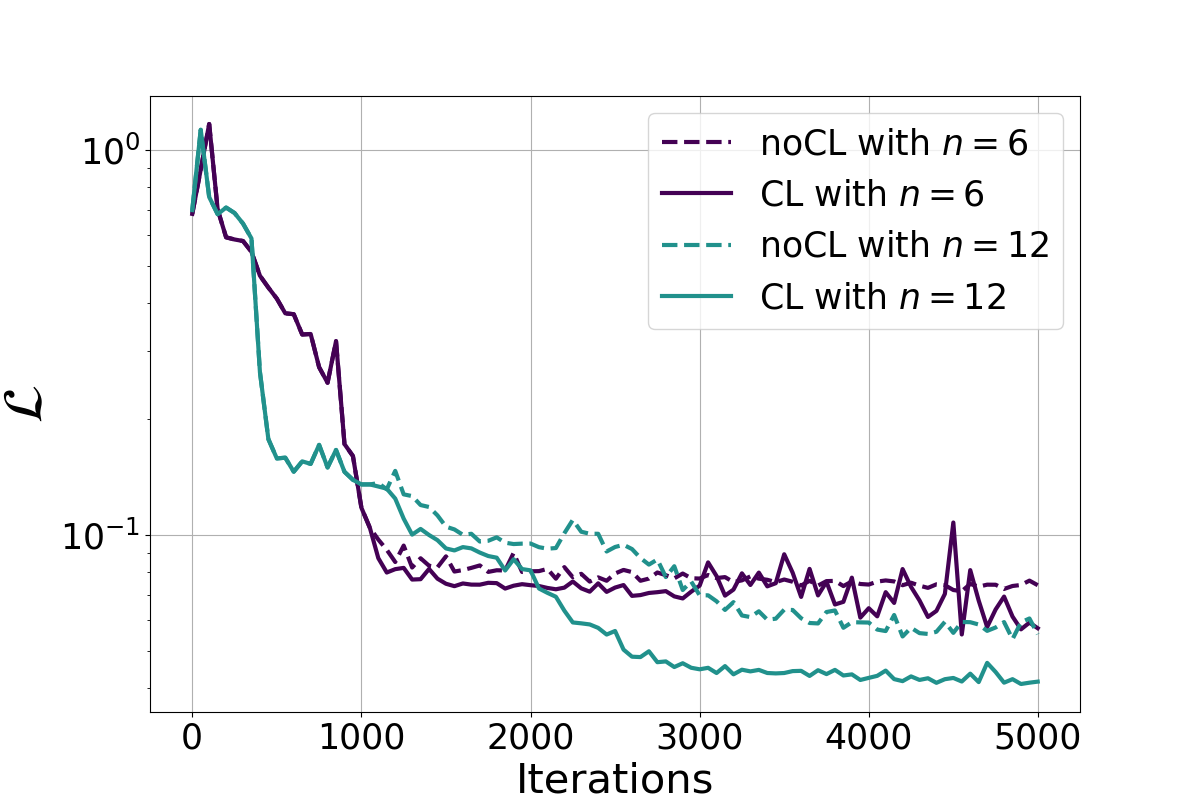}
\includegraphics[trim={0, 0 70, 60},clip,width=0.49\linewidth]{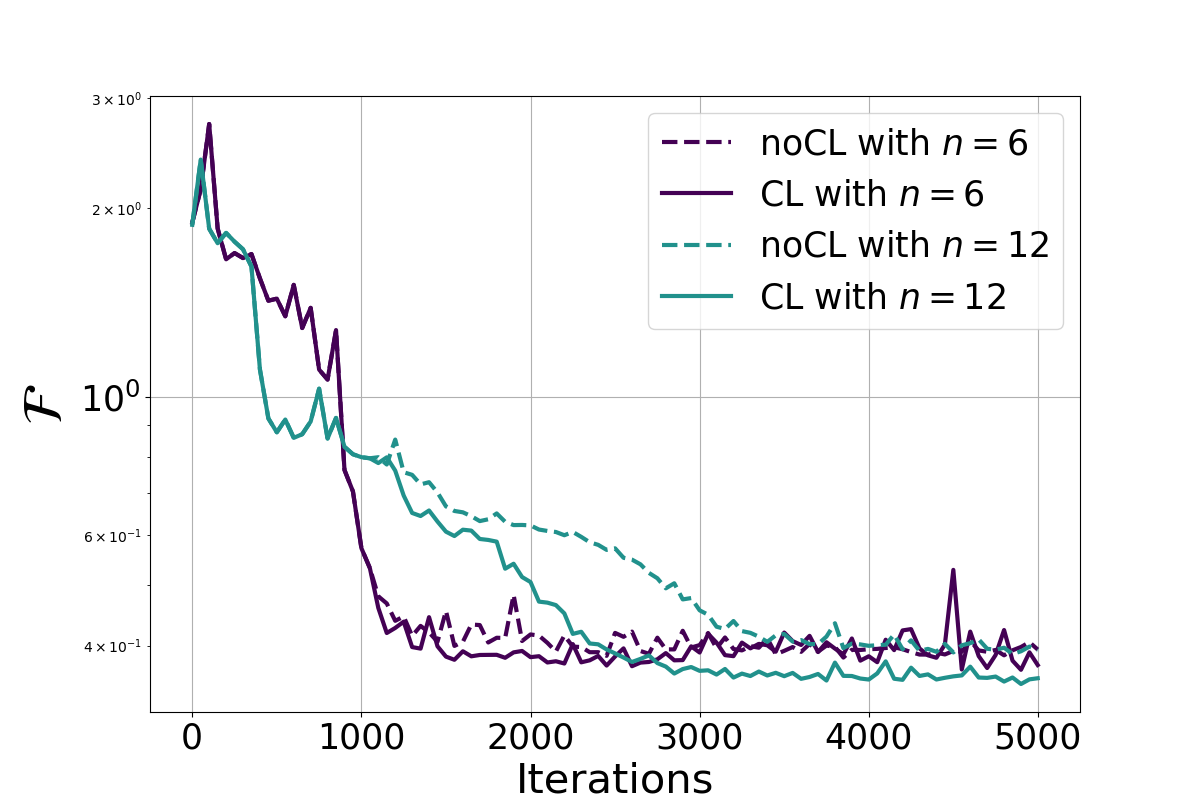}
\caption{\small Evolution of $\mathcal{L}$ (left) and $\mathcal{F}$ (right) during training. The panels represent policies trained with and without CL with solid and dashed lines, respectively. For $6$ (purple) and $12$ (blue) robots.}
\label{fig:metrics_evo_noiseless}
\end{figure}
With 12 robots, the benefits of CL become more pronounced: as congestion grows, early route selection and speed adjustments have greater long-term impact. CL helps the policy learn to anticipate and avoid crowded areas, making it more useful for the bigger team.
This effect is most apparent in the \textit{navigation} task, where multiple routes to the goal exist and long-term planning yields clear gains. While in \textit{passage} the same effect is present, it is reduced because the task forces every robot through the same narrow corridor, concentrating complexity in immediate collision avoidance and tight-space reorganization rather than long-term path planning. This characteristics are clearly shown in Fig. \ref{fig:qualitative}.
Interestingly, the number of robots that successfully complete the task increases with the level of noise in both scenarios, even when evaluated under the same noise level used during training. This is because the learned variability also acts as a lightweight exploration mechanism, since robots occasionally perturb their actions enough to escape local blockages. This controlled stochastic flexibility is learned during training and enables policies to recover from stalls, boosting the task success rate, even under noiseless conditions.

Figure \ref{fig:metrics_evo_noiseless} shows the evolution of the loss $\mathcal{L}$ and the Fr\'echet distance $\mathcal{F}$ for the \textit{navigation} task, computed in the same way as in Tab. \ref{tab:table_results}. Curriculum Learning shows increased sample-efficiency against the non-curriculum version, effectively reducing the number training iterations required to converge. This effect is, once again, especially strong with $12$ agents, where the congestion amplifies the importance of long-term decision making.

\begin{figure}
\includegraphics[trim={0, 0 70, 60},clip,width=0.49\linewidth]{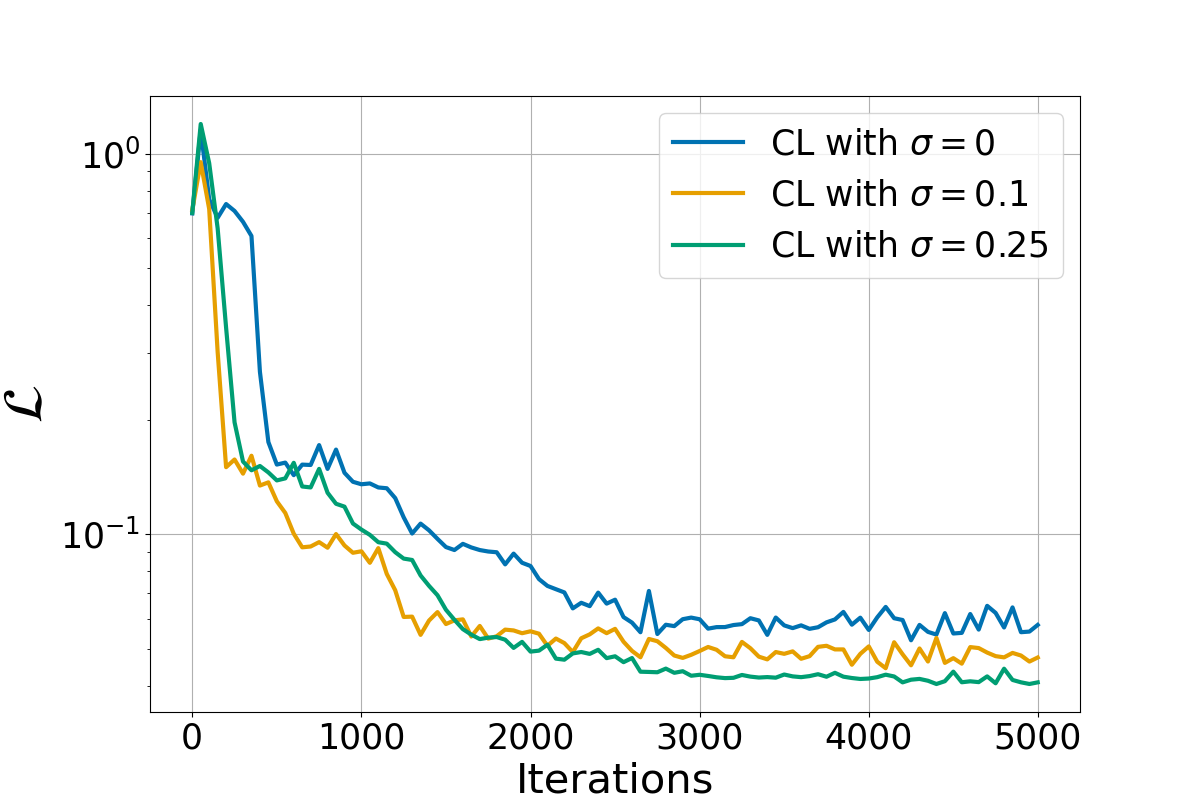}
\includegraphics[trim={0, 0 70, 60},clip,width=0.49\linewidth]{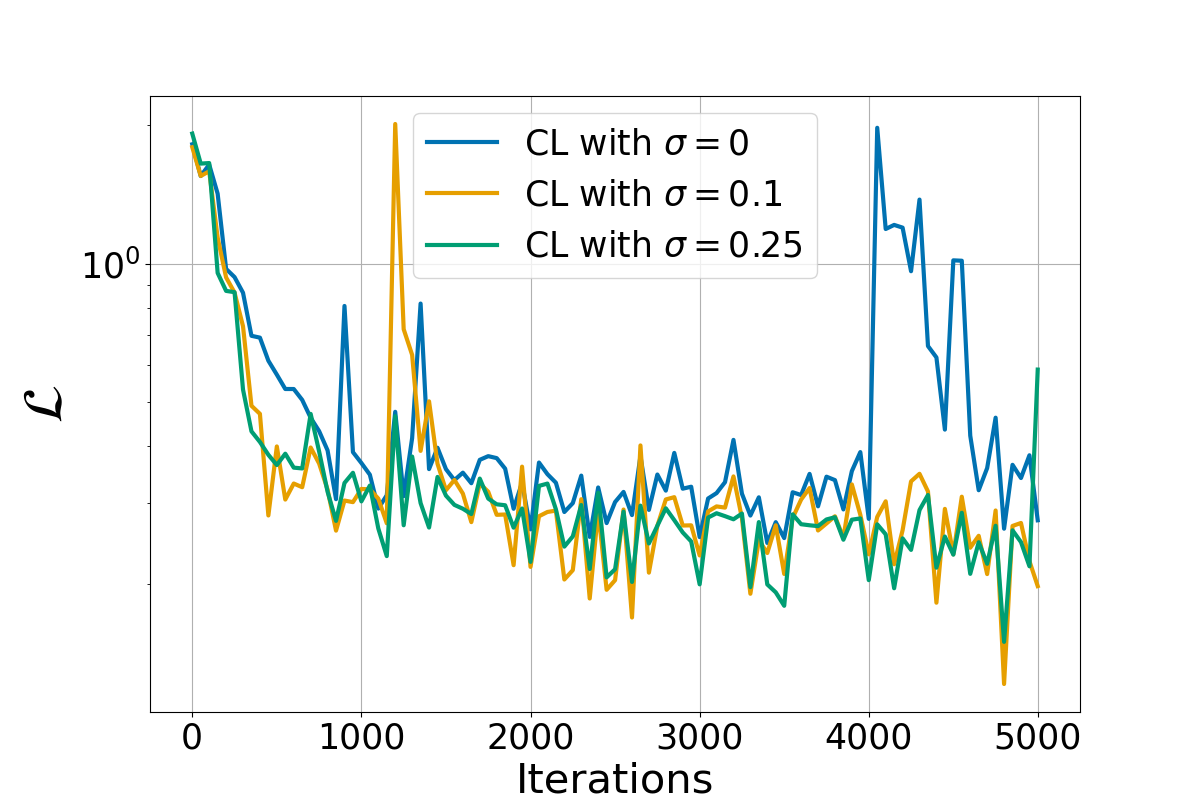}
\caption{\small Evolution of $\mathcal{L}$ in \textit{navigation} (left) and \textit{passage} (right)  with $n=12$. Compares policies trained with CL using $\sigma=\{0, 0.1,0.25\}$ (respectively: blue, yellow and green), but deployed with $\sigma=0.25$. }
\label{fig:metrics_evo_noisy}
\end{figure}
Fig.~\ref{fig:metrics_evo_noisy} shows the evolution of $\mathcal{L}$ during training for $12$ robots in both scenarios, all under high perceptual noise (\(\sigma=0.25\)). Unlike Tab.\ref{tab:table_results},
where loss converged to similar values, higher noises during training yield consistently lower losses, confirming that our egocentric observation estimation strengthens robustness to real-world uncertainty. This gain is especially clear in \textit{navigation}, where long-term planning remains stable. Although in \textit{passage}, loss remains lower with noise than without. The more constrained, collision-heavy nature of the \textit{passage} task and the destabilizing effect of CL introduce more variance in $\mathcal{L}$, making improvements less pronounced. Importantly, these CL-induced instabilities are not a problem when kept moderate, as the training progression naturally corrects them. In fact, noise injection itself appears to smoothen such fluctuations, showing synergy between both strategies in our method.


\section{Conclusions}\label{sec:conclusion}

In this paper, we have presented a Curriculum Learning method for imitation learning of distributed multi-robot policies.
We have considered a curriculum strategy with respect to the length of the expert trajectories to avoid stability problems related to long trajectories.
The paper has also dealt with the problem of transforming third-person observations into egocentric robot observations, enabling imitation learning from observations.
Both methods are integrated into a physics-informed learning framework, yielding explainable distributed control policies, implementable using robot noisy perception.
Extensive experiments on two representative tasks, under multiple team sizes and noise levels, confirm that (i) trajectory-length curriculum enhances sample efficiency and long-term fidelity, and (ii) egocentric observation modeling produces distributed policies resilient to sensor noise.
All together, our solution paves the way for deploying multi-robot imitation controllers in real-world settings, even when expert actions and egocentric sensor logs are unavailable.


\bibliographystyle{IEEEtran}
\bibliography{bibliography}

\end{document}